\documentclass[conference]{IEEEtran}
\IEEEoverridecommandlockouts

\usepackage{cite}
\usepackage{amsmath,amssymb,amsfonts}
\usepackage{graphicx}
\usepackage{textcomp}
\usepackage{xcolor}
\usepackage{url}
\usepackage{microtype}
\usepackage{balance}
\usepackage[hidelinks]{hyperref}

\usepackage{algorithm}
\usepackage{algpseudocode}

\usepackage{tikz}
\usetikzlibrary{arrows.meta,positioning,calc,shapes.geometric,fit,backgrounds}

\usepackage{placeins}
\usepackage{array}
\usepackage{booktabs}
\usepackage{multirow}
\usepackage{tabularx}
\usepackage{ragged2e}

\newcolumntype{L}[1]{>{\RaggedRight\arraybackslash}p{#1}}
\newcolumntype{Y}{>{\RaggedRight\arraybackslash}X}

\def\BibTeX{{\rm B\kern-.05em{\sc i\kern-.025em b}\kern-.08em
    T\kern-.1667em\lower.7ex\hbox{E}\kern-.125emX}}

\definecolor{boxfill}{RGB}{226,238,250}
\definecolor{boxedge}{RGB}{42,91,145}
\definecolor{accentfill}{RGB}{255,235,204}
\definecolor{accentedge}{RGB}{204,112,0}

\tikzset{
procbox/.style={
rectangle,
rounded corners=2pt,
draw=boxedge,
thick,
fill=boxfill,
align=center,
inner sep=4pt,
minimum height=10mm,
minimum width=36mm,
text width=33mm,
font=\scriptsize
},
databox/.style={
rectangle,
draw=boxedge,
thick,
fill=boxfill,
align=center,
inner sep=4pt,
minimum height=8mm,
minimum width=27mm,
text width=24mm,
font=\scriptsize
},
accentbox/.style={
rectangle,
rounded corners=2pt,
draw=accentedge,
very thick,
fill=accentfill,
align=center,
inner sep=4pt,
minimum height=10mm,
minimum width=36mm,
text width=33mm,
font=\scriptsize
},
flow/.style={
-{Stealth[length=2.4mm,width=1.6mm]},
thick,
draw=boxedge
},
edgelbl/.style={
font=\scriptsize,
midway,
fill=white,
inner sep=1.2pt,
rounded corners=1pt
}
}

\begin{document}
\title{GIST-CMTF: Goal-State Inference for Causal Minimal Tool Filtering in LLM Agents}

\author{\IEEEauthorblockN{Rahul Suresh Babu}
\IEEEauthorblockA{\textit{Independent Researcher} \\
United States of America \\
rahulsb@bu.edu}
\and
\IEEEauthorblockN{Rohit Shukla}
\IEEEauthorblockA{\textit{Independent Researcher} \\
United States of America \\
rohshukla@cs.stonybrook.edu}
}

\maketitle

\begin{abstract}
Tool-augmented LLM agents rely on runtime filtering to decide which tools should be visible at each step. Causal Minimal Tool Filtering (CMTF) reduces tool-choice confusion by exposing only the next causally necessary tool frontier, but it assumes that the user request has already been mapped to a symbolic goal state. In practice, requests such as ``handle my appointment'' or ``take care of this email'' may correspond to multiple possible goals. This creates wrong-goal execution, where an agent follows a valid causal tool path for an unintended objective. We introduce GIST-CMTF, a goal-state inference layer that predicts candidate symbolic goals over the same state-transition vocabulary used by CMTF, estimates ambiguity, and either applies CMTF or exposes clarification as a causal action that produces missing goal or state variables. We evaluate GIST-CMTF across seven model backends, six filtering methods, and 120 controlled tool-use tasks. GIST-CMTF achieves 97.0\% task success, compared with 80.1\% for top-goal CMTF and 82.9\% for semantic-goal CMTF. It reduces wrong-goal execution from 19.4\% under top-goal CMTF to 2.5\%, while preserving the one-tool exposure of causal filtering and using substantially fewer tokens than all-tools exposure. These results suggest that reliable tool-augmented agents should validate goal state, not only tool relevance, before exposing external actions.
\end{abstract}

\begin{IEEEkeywords}
Tool-augmented LLM agents, goal-state inference, causal tool filtering, Causal Minimal Tool Filtering, intent disambiguation, clarification, wrong-goal execution, symbolic task state, preconditions and effects, function calling, tool selection, LLM reliability, multi-step tool use, agent orchestration
\end{IEEEkeywords}

\section{Introduction}

Tool access has become a central mechanism for extending large language models (LLMs) beyond text generation. Modern LLM agents can interleave reasoning with external actions, invoke APIs, search information sources, operate over files, update calendars, draft emails, execute code, and interact with structured systems \cite{yao2023react,schick2023toolformer,qin2024toolllm,li2023apibank}. As tool ecosystems grow, however, agents face a runtime interface problem: before the model can use a tool correctly, the system must decide which tools should be visible at the current step. Function-calling benchmarks and tool-use evaluations measure whether models can select APIs, construct arguments, and complete multi-step tasks, but they often assume that the visible tool interface has already been defined \cite{patil2025bfcl,liu2024agentbench}.

Recent work on Causal Minimal Tool Filtering (CMTF) addresses this tool-exposure problem by treating tool visibility as a causal state-transition problem rather than a semantic relevance problem \cite{babu2026toolchoiceconfusion}. In CMTF, each tool is represented by a lightweight contract consisting of preconditions, effects, optional cost, and optional risk. Given a current symbolic task state and a target goal state, CMTF identifies a minimal causal path and exposes only the next causally necessary frontier. This reduces ToolChoiceConfusion: failures caused by exposing tools that are semantically plausible but unnecessary, premature, or non-goal-directed at the current step. Recent contract-learning work studies how such precondition-effect contracts can be inferred, but even accurate contracts still require a target goal state \cite{babu2026contract2tool}.

This requirement exposes a critical upstream assumption: the user request has already been mapped to a well-defined symbolic goal. In practical agent workflows, this assumption is often unrealistic. Users issue requests that are ambiguous, underspecified, or compatible with multiple possible goals. For example, ``handle my dentist appointment'' could mean finding the appointment, summarizing it, rescheduling it, canceling it, or asking the user what they want done. Similarly, ``take care of this email'' could mean summarizing it, drafting a reply, sending a reply, archiving it, deleting it, or flagging it for later. A causal filter can only expose the right next tool if it is optimizing toward the right goal. Thus, even a perfect causal tool filter can fail if it is given the wrong goal.

This paper studies goal-state inference as the missing upstream layer for causal tool filtering. We identify \emph{wrong-goal execution} as a distinct failure mode in tool-augmented agents. In wrong-goal execution, the system may expose and execute a causally valid sequence of tools, but for a goal the user did not intend. This differs from ordinary wrong-tool selection: the tool path may be internally valid relative to the inferred goal, yet externally incorrect relative to the user's intended objective. A system that maps ``take care of this email'' directly to an email-deletion goal, for instance, may execute a valid causal path to deletion while violating the user's actual intent.

We introduce \textbf{GIST-CMTF}, a goal-state inference layer for Causal Minimal Tool Filtering. GIST-CMTF formulates goal inference as structured prediction over the same symbolic state-transition vocabulary used by CMTF. Given a natural-language request, current task state, and tool contract graph, GIST-CMTF predicts candidate symbolic goal states, estimates goal ambiguity, and decides whether to apply CMTF or expose a clarification action. The goal is not merely to classify the user's intent into a natural-language label, but to infer the symbolic end state that downstream causal filtering should optimize toward.

A central design choice in GIST-CMTF is to model clarification as a causal action. When the user request is ambiguous or missing necessary goal parameters, the correct next step may not be an external API call. It may be a clarification question that produces a missing goal variable, entity variable, or confirmation variable. For example, a clarification action may transform an ambiguous-goal state into a goal-specified state. This allows clarification to be represented within the same precondition-effect framework as external tools, rather than being treated as an ad hoc fallback outside the agent's causal model.

This framing extends causal tool filtering from tool selection to goal selection. Prior CMTF work asks: given a current state and a goal, which tool should be exposed next? GIST-CMTF asks an earlier question: is the goal sufficiently specified to expose a downstream tool path at all? If the inferred goal is confident and sufficiently specified, GIST-CMTF passes that goal to CMTF. If the goal is ambiguous, missing required variables, or would commit the agent to a risky or irreversible path without sufficient intent evidence, GIST-CMTF exposes a clarification action instead.

We evaluate GIST-CMTF on controlled multi-step tool-use tasks containing explicit, ambiguous, missing-variable, and clarification-required requests across workflow domains such as calendar, email, files, contacts, and authorization. We measure both goal-layer behavior and downstream agent behavior, including task success, wrong-goal execution, clarification behavior, visible-tool exposure, and token cost. Across seven model backends, six filtering methods, and 120 controlled tool-use tasks, GIST-CMTF achieves 97.0\% task success, compared with 80.1\% for top-goal CMTF and 82.9\% for semantic-goal CMTF. It reduces wrong-goal execution from 19.4\% under top-goal CMTF and 16.7\% under semantic-goal CMTF to 2.5\%, while preserving the one-tool exposure of causal filtering.

This paper makes the following contributions:
\begin{enumerate}
\item We formulate goal-state inference as a missing upstream problem in causal tool filtering for LLM agents.
\item We identify wrong-goal execution as a distinct failure mode, where an agent follows a causally valid tool path for an unintended goal.
\item We introduce GIST-CMTF, a framework that predicts symbolic goal states over the same state-transition vocabulary used by CMTF and uses clarification as a causal action when the goal is ambiguous or underspecified.
\item We evaluate GIST-CMTF across explicit, ambiguous, missing-variable, and clarification-required tool-use tasks, measuring both goal-layer behavior and downstream agent reliability.
\end{enumerate}

Overall, GIST-CMTF suggests that reliable tool use requires more than selecting the right tool for the current state. Before exposing external actions, an agent must first validate which symbolic goal it is pursuing.

\section{Background}

\subsection{Tool-Augmented LLM Agents}

Tool use has become a central mechanism for extending large language models beyond text generation. Agentic systems can interleave reasoning and acting, invoke external APIs, retrieve information, operate over files, update calendars, draft emails, execute code, and interact with structured services \cite{yao2023react,schick2023toolformer,qin2024toolllm}. This capability has motivated benchmarks for function calling and tool-augmented dialogue, including evaluations of API selection, argument construction, multi-step tool use, and agent task completion \cite{li2023apibank,patil2025bfcl,liu2024agentbench}. These works establish tool use as a core capability for LLM agents, but they often assume that the visible tool interface has already been constructed. Reliability-oriented orchestration work further studies how agent runtimes can detect failures, apply bounded recovery actions, verify recovered trajectories, and record observability traces in tool-augmented LLM systems \cite{babu2026selfhealing}.

\subsection{Tool Exposure and Causal Minimal Tool Filtering}

As tool libraries grow, the visible tool menu becomes an important runtime control surface. Exposing all available tools can increase context cost and selection burden, while relevance-based retrieval may expose tools that are semantically related to the user request but inappropriate for the current step. Causal Minimal Tool Filtering (CMTF) addresses this problem by representing each tool as a lightweight state transition with preconditions and effects \cite{babu2026toolchoiceconfusion}. Given a current symbolic state $s_t$, a goal state $g$, and a tool library $T$, CMTF constructs a dependency graph and exposes only the next causally necessary frontier. This shifts tool filtering from semantic relevance to causal sufficiency: a tool should be visible when its effects help advance the current state toward the goal. Related work on capability minimization extends this least-privilege view by treating risk-aware causal gating as a safety primitive for limiting agent capabilities before external actions are exposed \cite{iyer2026capabilityminimization}.

\subsection{Tool Contracts and State-Transition Semantics}

CMTF relies on lightweight tool contracts. A tool $t_i$ can be represented as
\begin{equation}
t_i = (d_i, R_i, E_i, c_i, \rho_i),
\end{equation}
where $d_i$ is a natural-language description, $R_i$ is the set of required state variables, $E_i$ is the set of produced state variables, $c_i$ is an optional cost, and $\rho_i$ is an optional risk label. This abstraction is related to classical planning formalisms such as STRIPS and PDDL, which represent actions using preconditions and effects \cite{fikes1971strips,mcdermott1998pddl}. In tool-augmented LLM agents, however, the objective is not to replace the model with a symbolic planner, but to shape the action interface exposed to the model at each step.

Recent contract-learning work studies how these preconditions, effects, and risk annotations can be inferred from tool names, descriptions, schemas, documentation, and execution traces \cite{babu2026contract2tool}. This reduces the burden of manually writing contracts for large or changing tool ecosystems. However, even with accurate contracts, causal filtering still requires a target goal state.

\subsection{Goal-State Inference as the Missing Upstream Layer}

Existing causal filtering methods typically assume that the user's request has already been mapped to a symbolic goal state. In practice, this assumption may fail. A request such as ``handle my appointment'' may correspond to finding, summarizing, rescheduling, canceling, or asking for clarification about an event. Similarly, ``take care of this email'' may imply summarizing, drafting, sending, archiving, deleting, or flagging the message. These alternatives correspond to different symbolic goals and therefore different causal tool paths.

This paper studies goal-state inference as the missing upstream layer before causal tool filtering. The central challenge is not only to identify a natural-language intent label, but to infer the symbolic end state that downstream CMTF should optimize toward. When the goal is ambiguous or underspecified, the correct next action may be clarification rather than external tool execution. GIST-CMTF therefore extends causal filtering from selecting the next tool given a goal to first validating whether the goal itself is sufficiently specified.

\section{Problem Formulation}

We study goal-state inference for tool-augmented LLM agents that use causal tool filtering. Let $T = \{t_1, t_2, \ldots, t_n\}$ denote a tool library and let $X$ denote the vocabulary of symbolic task-state variables. Each tool $t_i \in T$ is represented by a lightweight contract:
\begin{equation}
t_i = (d_i, R_i, E_i, c_i, \rho_i),
\end{equation}
where $d_i$ is a natural-language description, $R_i \subseteq X$ is the set of required state variables, $E_i \subseteq X$ is the set of produced state variables, $c_i$ is an optional cost, and $\rho_i$ is an optional risk label. This precondition-effect representation follows the causal tool-filtering view introduced in prior CMTF work and is compatible with learned-contract approaches.

At runtime, the agent observes a natural-language user request $q$, a current symbolic task state $s_t \subseteq X$, and the available tool library $T$. Standard CMTF assumes that the target goal state $g \subseteq X$ is known. Under that assumption, the filtering method selects a visible tool set:
\begin{equation}
V_t = F(s_t, g, T),
\end{equation}
where $F$ exposes the next causally sufficient tool frontier for advancing from $s_t$ toward $g$.

In realistic settings, however, the goal state is not directly observed. The same request may correspond to multiple possible symbolic goals. For example, the request ``handle my appointment'' may be compatible with:
\begin{align}
g_1 &= \{event\_details\_found\}, \\
g_2 &= \{event\_summarized\}, \\
g_3 &= \{event\_updated\}, \\
g_4 &= \{event\_deleted\}.
\end{align}
These goals imply different causal tool paths and different exposure decisions.

We therefore define a goal-state inference module $H$ that maps the user request, current state, and tool-contract context to a ranked set of candidate goals:
\begin{equation}
H(q, s_t, T) = \{(g_1, p_1), (g_2, p_2), \ldots, (g_k, p_k)\},
\end{equation}
where each $g_i \subseteq X$ is a candidate symbolic goal state and $p_i$ is a confidence score. Let $g^\star$ denote the highest-confidence candidate goal:
\begin{equation}
g^\star = \arg\max_{g_i} p_i.
\end{equation}

The system must decide whether $g^\star$ is sufficiently specified to support downstream tool filtering. We define a goal-aware filtering policy:
\begin{equation}
\Pi(q, s_t, T) =
\begin{cases}
F(s_t, g^\star, T), & \text{if } A(q, s_t, G_q) = 0, \\
\{a_{clarify}\}, & \text{if } A(q, s_t, G_q) = 1,
\end{cases}
\end{equation}
where $G_q = H(q, s_t, T)$ is the candidate goal set, $A$ is an ambiguity detector, and $a_{clarify}$ is a clarification action.

We model clarification as a causal action rather than an external fallback. A clarification action has its own preconditions and effects:
\begin{equation}
a_{clarify} = (d_c, R_c, E_c, c_c, \rho_c),
\end{equation}
where $R_c$ may include variables such as $\{ambiguous\_goal\}$ or $\{missing\_entity\}$, and $E_c$ may include variables such as $\{goal\_specified\}$, $\{entity\_identified\}$, or $\{permission\_confirmed\}$. This allows clarification to be represented within the same state-transition framework as external tools.

We define \emph{wrong-goal execution} as the failure mode in which the agent executes a causally valid tool sequence for an inferred goal $\hat{g}$ that does not match the intended goal $g^{true}$. Formally, wrong-goal execution occurs when:
\begin{equation}
\hat{g} \neq g^{true}
\end{equation}
and the agent reaches $\hat{g}$ through a valid causal path. This differs from wrong-tool selection: the tool sequence may be valid relative to $\hat{g}$, but incorrect relative to the user's intended objective.

The objective of GIST-CMTF is to preserve the reliability and efficiency benefits of causal tool filtering while reducing wrong-goal execution. Ideally, the system should infer the intended symbolic goal when the request is clear, ask for clarification when the goal is ambiguous or underspecified, and avoid exposing downstream tools before the goal is sufficiently validated.

\section{Method: GIST-CMTF}

GIST-CMTF extends Causal Minimal Tool Filtering with an upstream goal-state inference layer. Standard CMTF assumes that the target symbolic goal state is known before filtering begins. GIST-CMTF relaxes this assumption by first inferring candidate goal states from the user request and then deciding whether the request is sufficiently specified to expose a downstream tool frontier. If the goal is clear, GIST-CMTF applies CMTF using the inferred goal. If the goal is ambiguous or underspecified, it exposes a clarification action instead.

\subsection{Overview}
Figure~\ref{fig:gist-architecture} illustrates the overall GIST-CMTF execution flow. Goal-state inference precedes causal tool filtering, allowing the system to determine whether clarification is required before exposing downstream actions. This separates goal validation from tool selection and prevents the system from committing to an incorrect symbolic objective before the user intent is sufficiently specified.

Given a user request $q$, current symbolic state $s_t$, and tool library $T$, GIST-CMTF proceeds in four stages:
\begin{enumerate}
    \item generate candidate symbolic goal states;
    \item estimate goal confidence and ambiguity;
    \item expose a clarification action when the goal is underspecified;
    \item otherwise apply CMTF using the selected goal.
\end{enumerate}

This design separates goal selection from tool selection. The goal-inference layer decides what end state the agent should pursue, while CMTF decides which tool should be visible next once that goal is accepted.

\definecolor{boxfill}{RGB}{232,242,255}
\definecolor{boxedge}{RGB}{35,95,155}

\begin{figure*}[t]
\centering
\resizebox{0.96\textwidth}{!}{%
\begin{tikzpicture}[
node distance=0.95cm and 0.75cm,
box/.style={
rectangle,
rounded corners=2pt,
draw=boxedge,
thick,
fill=boxfill,
align=center,
inner sep=4pt,
minimum height=10mm,
minimum width=30mm,
text width=28mm,
font=\scriptsize
},
arrow/.style={
-{Stealth[length=2.2mm,width=1.5mm]},
thick,
draw=boxedge
},
lbl/.style={
font=\scriptsize,
fill=white,
inner sep=1pt
}
]

\node[box] (nreq) {User request and current state};
\node[box, right=of nreq] (ngoal) {Goal-state inference};
\node[box, right=of ngoal] (ncheck) {Ambiguity and missing-variable check};

\node[box, above right=0.80cm and 0.70cm of ncheck] (nclarify)
{Clarification action};

\node[box, below right=0.80cm and 0.70cm of ncheck] (nfilter)
{CMTF with accepted goal};

\node[box, right=of nfilter] (nfrontier)
{Minimal causal tool frontier};

\node[box, right=of nfrontier] (naction)
{LLM action};

\node[box, right=of naction] (nstate)
{Symbolic state update};

\draw[arrow] (nreq) -- (ngoal);
\draw[arrow] (ngoal) -- (ncheck);

\draw[arrow] (ncheck.north east) -- node[lbl, above left]{ambiguous / underspecified} (nclarify.south west);
\draw[arrow] (ncheck.south east) -- node[lbl, below left]{goal accepted} (nfilter.north west);

\draw[arrow] (nclarify.east) -| (nstate.north);

\draw[arrow] (nfilter) -- (nfrontier);
\draw[arrow] (nfrontier) -- (naction);
\draw[arrow] (naction) -- (nstate);

\end{tikzpicture}
}
\caption{Overview of GIST-CMTF. The system first infers candidate symbolic goal states, checks whether the goal is ambiguous or underspecified, and either exposes a clarification action or applies CMTF using the accepted goal.}
\label{fig:gist-architecture}
\end{figure*}
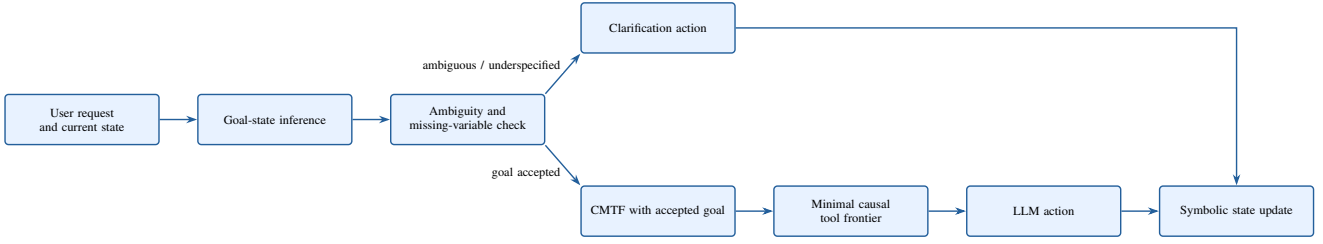

\subsection{Candidate Goal Generation}

The first stage maps the natural-language request $q$ into a ranked set of candidate symbolic goal states. Let $X$ be the state-variable vocabulary used by the tool contracts. GIST-CMTF predicts:
\begin{equation}
G_q = \{(g_1, p_1), (g_2, p_2), \ldots, (g_k, p_k)\},
\end{equation}
where each $g_i \subseteq X$ is a candidate goal and $p_i$ is its confidence score.

The candidate goals are drawn from the same symbolic vocabulary used by downstream causal filtering. This is important because GIST-CMTF does not predict only a natural-language intent label; it predicts the symbolic end state that determines the causal path over the tool graph. For example, the request ``take care of this email'' may yield candidate goals such as:
\begin{align}
g_1 &= \{email\_summarized\}, \\
g_2 &= \{draft\_created\}, \\
g_3 &= \{email\_sent\}, \\
g_4 &= \{email\_archived\}.
\end{align}

\subsection{Goal Confidence and Ambiguity Detection}

After candidate generation, GIST-CMTF determines whether the top goal is sufficiently reliable for downstream tool exposure. Let $g^\star$ denote the highest-confidence goal:
\begin{equation}
g^\star = \arg\max_{g_i \in G_q} p_i.
\end{equation}

The ambiguity detector considers both confidence and structural signals. A request may be treated as ambiguous when:
\begin{itemize}
    \item the top candidate has low confidence;
    \item multiple candidate goals have similar confidence;
    \item required goal parameters are missing;
    \item the request contains vague action verbs such as ``handle,'' ``fix,'' or ``take care of'';
    \item accepting the goal would commit the agent to a write, send, delete, share, or otherwise externally visible action.
\end{itemize}

In the simplest implementation, ambiguity can be decided using a confidence threshold $\tau$ and a margin threshold $\delta$:
\begin{equation}
A(q, s_t, G_q) =
\mathbb{I}\left[p^\star < \tau \ \lor \ (p^\star - p^{(2)}) < \delta\right],
\end{equation}
where $p^\star$ is the top confidence score and $p^{(2)}$ is the second-highest confidence score. More generally, $A$ may incorporate missing-variable checks, risk-sensitive rules, and learned ambiguity classifiers.

\subsection{Clarification as a Causal Action}

If the goal is ambiguous, GIST-CMTF exposes a clarification action instead of a downstream external tool. Clarification is represented as a tool-like causal action:
\begin{equation}
a_{clarify} = (d_c, R_c, E_c, c_c, \rho_c),
\end{equation}
where $R_c$ contains the ambiguity or missing-information condition and $E_c$ contains the state variable produced by clarification.

For example:

\begin{align}
a_g &: \{ambiguous\_goal\} \rightarrow \{goal\_specified\}, \\
a_e &: \{missing\_entity\} \rightarrow \{entity\_identified\}, \\
a_p &: \{missing\_permission\} \rightarrow \{permission\_confirmed\}.
\end{align}

Here, $a_g$, $a_e$, and $a_p$ denote goal, entity, and permission clarification actions, respectively.

This representation keeps clarification inside the same precondition-effect framework as external tool use. The system does not treat clarification as an ad hoc fallback; it treats clarification as the causally appropriate next action when the goal or required state is underspecified.

\subsection{Goal-Aware Causal Filtering}

If the goal is accepted, GIST-CMTF applies CMTF using $g^\star$. The visible tool set is:
\begin{equation}
V_t = F(s_t, g^\star, T),
\end{equation}
where $F$ is the CMTF filtering function that returns the next causally sufficient frontier.

The overall policy is:
\begin{equation}
\Pi(q, s_t, T) =
\begin{cases}
F(s_t, g^\star, T), & \text{if } A(q, s_t, G_q) = 0, \\
\{a_{clarify}\}, & \text{if } A(q, s_t, G_q) = 1.
\end{cases}
\end{equation}

After a clarification action is executed, the user's response updates the task state and goal state. GIST-CMTF then reruns goal inference or directly applies CMTF if the clarification resolves the ambiguity.

\subsection{Algorithm}

Algorithm~\ref{alg:gist-cmtf} summarizes GIST-CMTF.

\begin{algorithm}[t]
\footnotesize
\caption{GIST-CMTF}
\label{alg:gist-cmtf}
\begin{algorithmic}[1]
\Require User request $q$, current state $s_t$, tool library $T$, goal generator $H$, ambiguity detector $A$, CMTF filter $F$
\Ensure Visible action set $V_t$
\State $G_q \gets H(q, s_t, T)$
\State $g^\star \gets \arg\max_{g_i \in G_q} p_i$
\If{$A(q, s_t, G_q) = 1$}
    \State \Return $\{a_{clarify}\}$
\Else
    \State $V_t \gets F(s_t, g^\star, T)$
    \State \Return $V_t$
\EndIf
\end{algorithmic}
\end{algorithm}

\subsection{Failure Modes Addressed}

GIST-CMTF targets a failure mode that is distinct from wrong-tool selection. In wrong-tool selection, the agent chooses an incorrect tool relative to a known goal. In wrong-goal execution, the agent may choose tools that are causally valid relative to an inferred goal, but the inferred goal itself is not the user's intended objective. By validating the goal before exposing the downstream tool frontier, GIST-CMTF reduces the chance that the agent executes a coherent but unintended workflow.

\subsection{Implementation Variants}

We consider three implementation variants:
\begin{itemize}
    \item \textbf{Top-goal GIST-CMTF:} always use the highest-confidence inferred goal.
    \item \textbf{Thresholded GIST-CMTF:} use confidence and margin thresholds to decide whether to clarify.
    \item \textbf{Risk-sensitive GIST-CMTF:} require stronger goal confidence before exposing write, send, delete, share, or externally visible tool paths.
\end{itemize}

These variants allow us to separate the value of goal inference from the value of clarification and ambiguity-aware tool exposure.

\section{Experimental Setup}

We evaluate whether goal-state inference improves causal tool filtering under ambiguous and underspecified user requests. The evaluation is designed to isolate the effect of the upstream goal layer rather than introduce a new benchmark. We build on the controlled multi-step tool-use setting used in prior causal tool-filtering work, where tools have precondition-effect contracts, execution is deterministic, and task state is represented symbolically \cite{babu2026toolchoiceconfusion}. This allows us to measure whether failures arise from goal inference, tool exposure, or downstream tool selection.

\subsection{Evaluation Tasks}

We construct a focused evaluation suite across common workflow domains: calendar, email, files/documents, contacts, and authorization or confirmation workflows. Each task consists of a natural-language user request, an initial symbolic state, a set of candidate symbolic goals, an intended goal when one is specified, and a downstream gold tool path after the goal is known.

The evaluation includes four request classes. \textit{Explicit-goal requests} directly specify the intended goal, such as moving a calendar event or drafting an email reply. \textit{Ambiguous-goal requests} use underspecified verbs such as ``handle,'' ``take care of,'' or ``deal with,'' and are compatible with multiple goals. \textit{Missing-variable requests} specify an action but omit required information, such as the file to share, the intended recipient, or the permission level. \textit{Clarification-required requests} are cases where the correct next action is to ask the user for additional information before exposing downstream tools. Table~\ref{tab:benchmark-composition} summarizes the 120-task evaluation suite. Each task is evaluated under six filtering methods and seven model backends, yielding 5,040 total runs.

\begin{table}[t]
\centering
\caption{Benchmark composition by request type.}
\label{tab:benchmark-composition}
\small
\begin{tabular}{p{0.30\columnwidth}r p{0.50\columnwidth}}
\hline
Request type & Tasks & Purpose \\
\hline
Explicit goal & 40 & Clear objective; no clarification needed. \\
Ambiguous goal & 40 & Multiple plausible symbolic goals. \\
Missing variable & 30 & Required information is absent. \\
Clarification required & 10 & Clarification is the correct next action. \\
\hline
Total & 120 & Full evaluation suite. \\
\hline
\end{tabular}
\end{table}

\subsection{Compared Methods}

We compare GIST-CMTF against the following baselines:

\begin{itemize}
    \item \textbf{All tools:} exposes the full tool registry at each step.
    \item \textbf{State-aware filtering:} exposes tools whose preconditions are satisfied by the current symbolic state.
    \item \textbf{Gold-goal CMTF:} applies CMTF using the annotated intended goal. This serves as an upper-bound causal filtering baseline.
    \item \textbf{Top-goal CMTF:} infers a goal and always applies CMTF using the highest-confidence candidate, without clarification.
    \item \textbf{Semantic-goal CMTF:} selects the goal whose natural-language description is most similar to the user request, then applies CMTF.
    \item \textbf{GIST-CMTF:} infers candidate symbolic goals, estimates ambiguity, and either applies CMTF using the selected goal or exposes a clarification action.
\end{itemize}

This comparison separates three questions: whether causal filtering helps once the goal is known, whether naive goal inference is sufficient, and whether ambiguity-aware clarification improves downstream reliability.

\subsection{Execution Protocol}

Each run proceeds in a bounded step-by-step tool-use loop. At the first step, a method receives the user request, current symbolic state, and tool-contract context. Methods that require goal inference first select or infer a target goal. If GIST-CMTF detects that the goal is ambiguous or underspecified, it exposes a clarification action rather than a downstream external tool. Once the goal is specified, CMTF exposes the next causally necessary tool frontier.

At each step, the tool-calling model selects one visible action. The environment returns a deterministic mocked observation and updates the symbolic state according to the selected tool's effects. The run terminates when the intended goal is reached, an incorrect or invalid tool path is executed, a wrong-goal outcome is reached, no valid action is exposed, or the maximum step limit is exceeded.

\subsection{Metrics}

We report metrics that capture both goal-layer behavior and downstream tool-use outcomes.

Goal-layer metrics characterize how the filtering policy handles goal uncertainty:
\begin{itemize}
\item \textbf{Goal correctness:} whether the accepted goal matches the intended goal when a goal is selected.
\item \textbf{Ambiguity handling:} whether the method asks for clarification when the request is ambiguous or underspecified.
\item \textbf{Missing-variable handling:} whether the system asks for clarification when required entities, permissions, or goal parameters are absent.
\item \textbf{Unnecessary clarification:} whether clarification is requested for clear requests.
\end{itemize}

Downstream metrics measure agent behavior after goal inference and filtering:
\begin{itemize}
\item \textbf{Task success:} whether the intended goal is reached within the step limit.
\item \textbf{Wrong-goal execution:} whether the agent reaches a causally valid but unintended goal.
\item \textbf{Premature tool exposure:} whether downstream tools are exposed before the goal or required variables are specified.
\item \textbf{Wrong-tool calls:} number of selected tools that differ from the gold next tool for the intended goal.
\item \textbf{Tools exposed per step:} average visible action-set size.
\item \textbf{Trajectory length and token cost:} number of model-tool steps and total token usage per task.
\end{itemize}

\subsection{Analysis Plan}

The main analysis tests whether GIST-CMTF reduces wrong-goal execution without sacrificing the efficiency benefits of CMTF. We compare performance separately on explicit, ambiguous, missing-variable, and clarification-required requests. We also analyze the confidence threshold used by GIST-CMTF, measuring the tradeoff between unnecessary clarification and wrong-goal execution. Finally, we report whether clarification reduces premature exposure of write, send, delete, share, or other goal-committing actions.

The results are organized around four questions: (1) whether goal-state inference improves downstream task success, (2) whether ambiguity-aware clarification reduces wrong-goal execution, (3) how clarification affects tool exposure and token cost, and (4) whether the benefits generalize across model families.
\section{Results}

\begin{figure*}[t]
\centering
\includegraphics[width=\textwidth]{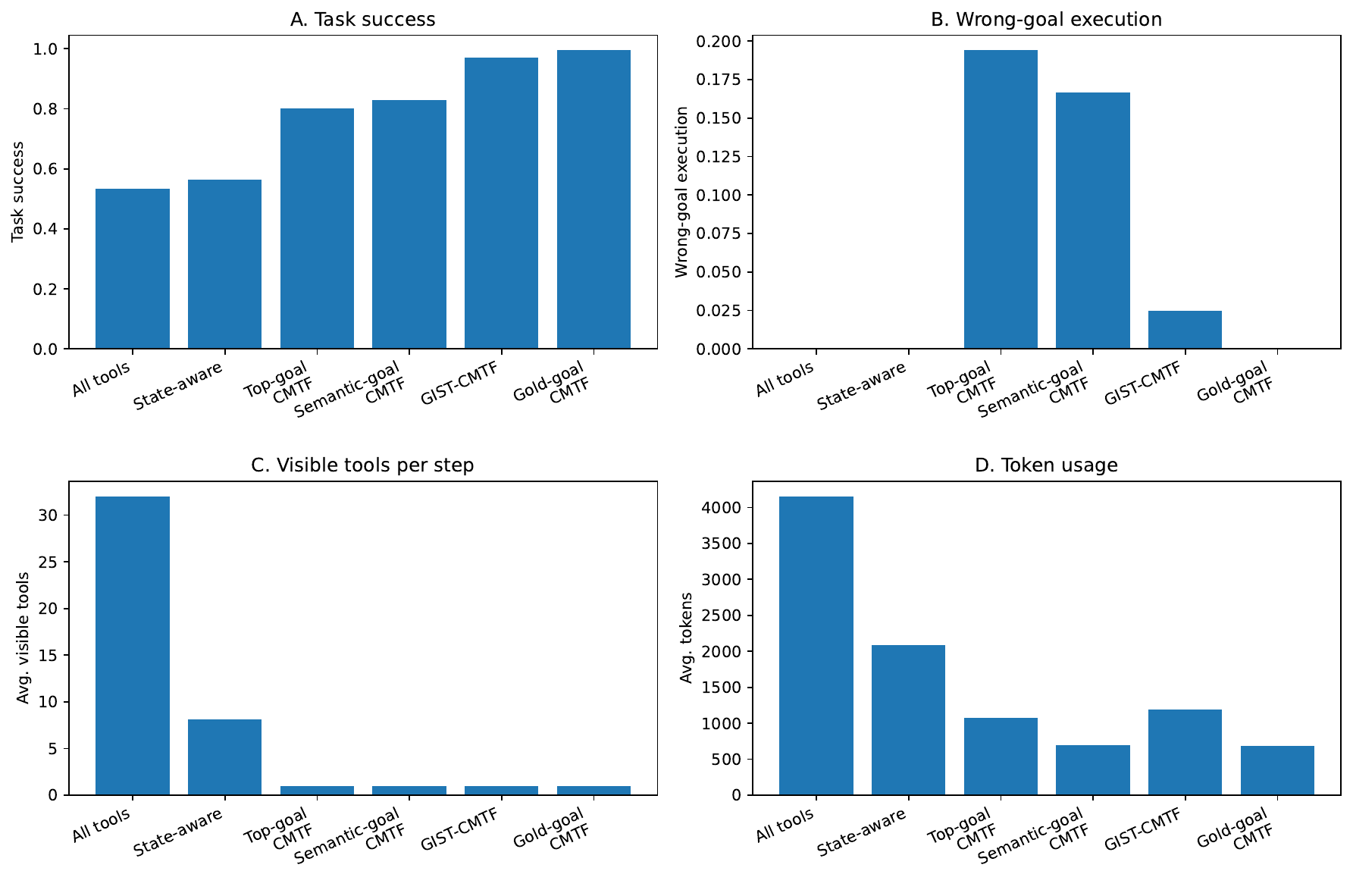}
\caption{Aggregate downstream performance across six filtering methods. GIST-CMTF substantially improves task success and reduces wrong-goal execution while preserving the one-tool exposure of causal filtering.}
\label{fig:main-results}
\end{figure*}

We evaluate six filtering methods across seven model backends and 120 tasks, producing 5,040 completed runs. The final dataset contains no duplicate model--method--task rows, no zero-token rows, and no remaining infrastructure-level provider errors. Table~\ref{tab:main-results} reports aggregate downstream performance, and Figure~\ref{fig:main-results} summarizes the main tradeoffs.

\subsection{Aggregate Downstream Performance}

\begin{table*}[t]
\centering
\caption{Aggregate downstream performance by filtering method. Higher task success is better; lower wrong-goal execution, visible tools, premature exposure, unnecessary clarification, and token cost are better.}
\label{tab:main-results}
\begin{tabular}{lrrrrrrrr}
\hline
Method & Success & Wrong-goal & Tools & Wrong tools & Prem. exp. & Clarif. & Unnec. clarif. & Tokens \\
\hline
All tools & 0.535 & 0.000 & 32.000 & 0.788 & 3.539 & 0.337 & 0.000 & 4152 \\
State-aware & 0.564 & 0.000 & 8.123 & 0.844 & 0.000 & 0.468 & 0.017 & 2084 \\
Top-goal CMTF & 0.801 & 0.194 & 1.000 & 0.642 & 0.000 & 0.331 & 0.000 & 1077 \\
Semantic-goal CMTF & 0.829 & 0.167 & 1.000 & 0.533 & 0.000 & 0.283 & 0.000 & 700 \\
GIST-CMTF & 0.970 & 0.025 & 1.000 & 1.051 & 0.000 & 0.660 & 0.056 & 1186 \\
Gold-goal CMTF & 0.995 & 0.000 & 1.000 & 0.332 & 0.000 & 0.167 & 0.000 & 689 \\
\hline
\end{tabular}
\end{table*}

GIST-CMTF achieves strong downstream performance without assuming access to the gold goal state. Across all runs, GIST-CMTF reaches 97.0\% task success, compared with 53.5\% for all-tools exposure, 56.4\% for state-aware filtering, 80.1\% for top-goal CMTF, and 82.9\% for semantic-goal CMTF. The oracle upper bound, gold-goal CMTF, reaches 99.5\% success. Thus, GIST-CMTF closes most of the gap between practical inferred-goal filtering and gold-goal causal filtering.

The strongest non-oracle comparison is against the two goal-inference baselines. Both top-goal CMTF and semantic-goal CMTF expose a minimal one-tool frontier after selecting an inferred goal, but they commit to that goal without ambiguity-aware validation. GIST-CMTF improves success by 16.9 percentage points over top-goal CMTF and 14.2 percentage points over semantic-goal CMTF, showing that the gains come from goal validation and clarification rather than from smaller tool menus alone.

\subsection{Wrong-Goal Execution}

Wrong-goal execution is the central failure mode targeted by GIST-CMTF. Top-goal CMTF has a wrong-goal execution rate of 19.4\%, and semantic-goal CMTF has a wrong-goal execution rate of 16.7\%. GIST-CMTF reduces this rate to 2.5\%, corresponding to an 87.1\% relative reduction versus top-goal CMTF and an 85.0\% relative reduction versus semantic-goal CMTF.

This result supports the main hypothesis of the paper: causal filtering is not sufficient when the target goal state is inferred incorrectly. Top-goal CMTF, semantic-goal CMTF, and GIST-CMTF all expose one tool per step on average, but only GIST-CMTF substantially reduces wrong-goal execution. The difference is therefore not simply a tool-exposure effect; it comes from validating whether the inferred symbolic goal is sufficiently specified before exposing downstream actions.

\subsection{Request-Type Breakdown and Clarification Behavior}

Table~\ref{tab:request-type} breaks performance down by request type. The largest gains occur on ambiguous-goal requests, where the user request is compatible with multiple symbolic goals. On these tasks, top-goal CMTF reaches 52.9\% success with 46.8\% wrong-goal execution, and semantic-goal CMTF reaches 50.0\% success with 50.0\% wrong-goal execution. GIST-CMTF reaches 97.5\% success and reduces wrong-goal execution to 2.1\%. This directly validates the goal-state inference layer: when the user request is ambiguous, clarification prevents the agent from executing a valid causal path for the wrong objective.

\begin{table*}[!t]
\centering
\caption{Performance by request type for key methods. GIST-CMTF provides the largest gains on ambiguous-goal requests, where naive goal inference frequently commits to the wrong symbolic goal.}
\label{tab:request-type}
\begin{tabular}{llrrrr}
\hline
Request type & Method & Success & Wrong-goal & Clarif. & Tokens \\
\hline
Ambiguous goal & All tools & 0.714 & 0.000 & 0.518 & 3371 \\
Ambiguous goal & Top-goal CMTF & 0.529 & 0.468 & 0.314 & 1015 \\
Ambiguous goal & Semantic-goal CMTF & 0.500 & 0.500 & 0.125 & 585 \\
Ambiguous goal & GIST-CMTF & 0.975 & 0.021 & 0.975 & 1239 \\
Ambiguous goal & Gold-goal CMTF & 0.993 & 0.000 & 0.000 & 608 \\
\hline
Clarification required & All tools & 0.943 & 0.000 & 0.057 & 2454 \\
Clarification required & Top-goal CMTF & 0.900 & 0.100 & 0.729 & 1015 \\
Clarification required & Semantic-goal CMTF & 1.000 & 0.000 & 0.900 & 717 \\
Clarification required & GIST-CMTF & 1.000 & 0.000 & 0.914 & 1066 \\
Clarification required & Gold-goal CMTF & 1.000 & 0.000 & 0.000 & 462 \\
\hline
Explicit goal & All tools & 0.300 & 0.000 & 0.000 & 5559 \\
Explicit goal & Top-goal CMTF & 0.904 & 0.086 & 0.000 & 1060 \\
Explicit goal & Semantic-goal CMTF & 0.989 & 0.000 & 0.000 & 690 \\
Explicit goal & GIST-CMTF & 0.946 & 0.050 & 0.168 & 1114 \\
Explicit goal & Gold-goal CMTF & 1.000 & 0.000 & 0.000 & 691 \\
\hline
Missing variable & All tools & 0.471 & 0.000 & 0.638 & 3882 \\
Missing variable & Top-goal CMTF & 0.995 & 0.005 & 0.662 & 1201 \\
Missing variable & Semantic-goal CMTF & 0.995 & 0.000 & 0.667 & 860 \\
Missing variable & GIST-CMTF & 0.986 & 0.005 & 0.810 & 1251 \\
Missing variable & Gold-goal CMTF & 0.990 & 0.000 & 0.667 & 870 \\
\hline
\end{tabular}
\end{table*}

On clarification-required tasks, GIST-CMTF reaches 100.0\% success with no wrong-goal execution, while requesting clarification in 91.4\% of runs. On missing-variable tasks, GIST-CMTF reaches 98.6\% success and clarifies in 81.0\% of runs, reflecting that the method often detects absent entities, permissions, or goal parameters before exposing downstream tools.

The main cost of GIST-CMTF is occasional over-clarification on clear requests. On explicit-goal tasks, GIST-CMTF reaches 94.6\% success, while semantic-goal CMTF reaches 98.9\% and gold-goal CMTF reaches 100.0\%. GIST-CMTF requests unnecessary clarification on 16.8\% of explicit-goal runs, and 5.6\% of all runs overall. This indicates a reliability--friction tradeoff: ambiguity-sensitive filtering substantially reduces wrong-goal execution, but can sometimes ask for clarification when the user goal was already clear.

\subsection{Tool Exposure and Token Cost}

GIST-CMTF preserves the minimal-exposure property of causal filtering. It exposes 1.0 visible tool per step on average, matching top-goal CMTF, semantic-goal CMTF, and gold-goal CMTF. In contrast, state-aware filtering exposes 8.1 tools per step, and all-tools exposure presents the full 32-tool registry.

Token usage follows the same pattern. GIST-CMTF uses 1,186 tokens per task on average, compared with 4,152 tokens for all-tools exposure and 2,084 tokens for state-aware filtering. GIST-CMTF is more expensive than semantic-goal CMTF and gold-goal CMTF because it performs ambiguity handling and clarification, but this additional cost buys a large reduction in wrong-goal execution and a substantial improvement in task success.

\subsection{Robustness Across Model Backends}

\begin{table*}[!t]
\centering
\caption{Task success by model and method. GIST-CMTF remains close to gold-goal CMTF across seven model backends.}
\label{tab:model-robustness}
\begin{tabular}{lrrrrrr}
\hline
Model & All tools & State-aware & Top-goal & Semantic-goal & GIST-CMTF & Gold-goal \\
\hline
Claude Opus 4.8 & 0.975 & 0.983 & 0.883 & 0.833 & 1.000 & 1.000 \\
Claude Sonnet 4.6 & 0.883 & 0.883 & 0.867 & 0.833 & 1.000 & 1.000 \\
Claude Haiku 4.5 & 0.292 & 0.400 & 0.775 & 0.833 & 0.983 & 1.000 \\
GPT-OSS-120B & 0.642 & 0.475 & 0.892 & 0.808 & 0.967 & 0.975 \\
Nova Premier & 0.383 & 0.475 & 0.708 & 0.825 & 0.975 & 0.992 \\
Nova 2 Lite & 0.258 & 0.358 & 0.683 & 0.833 & 0.892 & 1.000 \\
Nova Pro v1 & 0.308 & 0.375 & 0.800 & 0.833 & 0.975 & 1.000 \\
\hline
\end{tabular}
\end{table*}

Table~\ref{tab:model-robustness} reports task success by model and method. GIST-CMTF is consistently strong across all seven model backends. It reaches 100.0\% success with Claude Opus 4.8 and Claude Sonnet 4.6, 98.3\% with Claude Haiku 4.5, 96.7\% with GPT-OSS-120B, 97.5\% with Nova Premier, 89.2\% with Nova 2 Lite, and 97.5\% with Nova Pro. In every model family, GIST-CMTF substantially outperforms all-tools exposure and state-aware filtering, and it is consistently closer to gold-goal CMTF than the naive goal-inference baselines.

For the final seven-model aggregate, we use Nova Pro v1 in place of the unavailable Nova 2 Pro preview profile. We report the model identifier explicitly in Table~\ref{tab:model-robustness} for reproducibility.

\subsection{Error Analysis}

After cleaning infrastructure-corrupted rows, the final dataset contains no zero-token rows and no remaining provider-side infrastructure errors. The remaining errors are benchmark-level outcomes: 128 precondition failures, 24 tool-parse errors, 14 empty-text responses, 3 goal-parse errors, and 1 invalid-tool-selection error. These errors are retained because they reflect downstream agent behavior under the evaluated filtering policies rather than failed experiment execution.

\section{Discussion}

\subsection{Goal Validation as a Missing Layer in Tool Filtering}

The results show that reliable tool filtering is not only a tool-selection problem; it is also a goal-validation problem. CMTF is effective when the target goal is known, but practical user requests often arrive before that goal has been specified. In such cases, exposing a minimal causal frontier is not sufficient if the frontier is computed for the wrong inferred goal. GIST-CMTF addresses this upstream gap by validating whether the symbolic goal is clear enough before downstream tools are exposed.

\subsection{Wrong-Goal Execution Is Distinct from Wrong-Tool Selection}

Wrong-goal execution is more subtle than ordinary wrong-tool selection. In wrong-tool selection, the agent chooses an action that is incorrect relative to a known goal. In wrong-goal execution, the agent may follow a coherent and causally valid tool path, but for an unintended objective. This distinction explains why top-goal CMTF and semantic-goal CMTF can expose only one tool per step while still producing much higher wrong-goal execution than GIST-CMTF. The failure is not excessive tool exposure; it is premature commitment to the wrong symbolic goal.

\subsection{Clarification as a First-Class Causal Action}

A key design choice in GIST-CMTF is to treat clarification as a causal action rather than an external fallback. When the request is ambiguous, missing an entity, or missing a permission variable, the correct next action may be to ask the user for more information. Representing clarification with preconditions and effects allows it to participate in the same state-transition framework as external tools. This makes clarification a principled runtime action: it transforms an underspecified state into one where causal filtering can proceed safely.

\subsection{The Reliability--Friction Tradeoff}

GIST-CMTF improves reliability by reducing wrong-goal execution, but it introduces some additional friction. It asks more clarifying questions and uses more tokens than simpler one-tool baselines such as semantic-goal CMTF. This tradeoff is expected: ambiguity-sensitive systems must sometimes spend interaction cost to avoid committing to the wrong goal. The important practical implication is that clarification thresholds should depend on domain risk. For low-risk read-only actions, a system may tolerate more aggressive goal inference. For send, delete, share, update, or externally visible actions, stronger goal evidence and more conservative clarification are justified.

\subsection{Implications for Runtime Agent Mediation}

GIST-CMTF suggests that tool-augmented agents should mediate three decisions before acting: whether the goal is valid, whether the required state variables are present, and whether a candidate tool is causally appropriate for the accepted goal. This architecture shifts the runtime from tool retrieval toward goal-aware action mediation. Rather than asking only which tools are semantically relevant to the request, the agent runtime asks whether it should expose any downstream tool path at all. This is especially important for ambiguous or irreversible workflows, where a fluent but wrongly directed agent can be more harmful than one that asks for clarification.

\section{Limitations and Threats to Validity}

This study uses a controlled benchmark with synthetic tool-use tasks and mocked tool execution. This design allows us to isolate goal inference, clarification, tool exposure, and wrong-goal execution under known symbolic states and intended goals. However, real agent deployments may involve noisier user requests, incomplete state, changing APIs, authorization constraints, nondeterministic tool outputs, and irreversible side effects.

GIST-CMTF assumes that user requests, tool preconditions, tool effects, and candidate goals can be represented in a shared symbolic state vocabulary. This assumption is useful for causal filtering, but defining and maintaining the right vocabulary is itself a design challenge. Different domains may require different abstraction levels, and future work should study how these symbolic goal and state representations can be learned, updated, or transferred across tool ecosystems.

The current evaluation tests goal inference over a controlled candidate-goal set. This lets us measure wrong-goal execution precisely, but it does not fully address open-world goal discovery, where users may express goals not present in the candidate set. Similarly, clarification is modeled as an action that resolves a missing goal, entity, permission, or state variable. Real clarification dialogues may require multiple turns and may produce incomplete, ambiguous, or conflicting user responses.

Although GIST-CMTF reduces wrong-goal execution and premature commitment in the evaluated setting, we do not claim full production safety for high-risk tools. Actions such as sending, deleting, sharing, purchasing, or externally modifying state require additional safeguards, including authorization checks, policy enforcement, auditability, and human review where appropriate. Goal validation should be viewed as one layer in a broader runtime mediation stack.

Finally, while the results are consistent across seven model backends, model behavior may change with new versions, prompting strategies, tool-calling APIs, or deployment environments. We report token usage, but do not directly measure latency, monetary cost, or user satisfaction under clarification. Future work should evaluate GIST-CMTF with real users, real APIs, multi-turn clarification, larger dynamic tool registries, and risk-sensitive deployment policies.

\section{Conclusion}
\label{sec:conclusion}

This paper introduced GIST-CMTF, a goal-state inference layer for Causal Minimal Tool Filtering in tool-augmented LLM agents. The central motivation is that causal filtering can only expose the right next tool if the system is pursuing the right symbolic goal. In realistic requests, the intended goal may be ambiguous, underspecified, or missing required variables. GIST-CMTF addresses this upstream problem by inferring candidate symbolic goals, detecting ambiguity, and representing clarification as a causal action before downstream tools are exposed.

Across seven model backends, six filtering methods, and 120 tasks, GIST-CMTF achieved 97.0\% task success and reduced wrong-goal execution to 2.5\%, compared with 19.4\% under top-goal CMTF. It preserved the one-tool exposure of causal filtering while substantially improving reliability under ambiguous and underspecified requests. These results suggest that reliable tool-augmented agents need more than tool selection: they need runtime mediation of goal validity, state readiness, and causal tool exposure before acting.

\section*{Acknowledgment}
The authors thank colleagues for helpful feedback. This work was conducted in the authors' personal capacity. The views expressed in this paper are solely those of the authors and do not necessarily reflect the views of their employers. This work did not receive external funding. The authors declare no conflicts of interest.

\section*{Artifact Availability}

The benchmark, tool registry, filtering implementations, evaluation scripts, prompts, and analysis utilities used in this study are publicly available at: \url{https://github.com/R-Suresh/GIST-CMTF}. The repository contains task definitions, evaluation harnesses, and scripts required to reproduce the reported results.

\balance
\bibliographystyle{IEEEtran}
\bibliography{references}

\end{document}